\documentclass[runningheads]{llncs}

\usepackage{eccv} 
\usepackage{eccvabbrv}   % \eg \ie \etal \etc \cf

\usepackage[T1]{fontenc}
\usepackage{graphicx}
\usepackage{pgfplots}
\pgfplotsset{compat=1.18}
\usepackage[dvipsnames]{xcolor}
\usepackage{amsmath}
\usepackage{amsfonts}
\usepackage{amssymb}
\usepackage{multirow}
\usepackage{booktabs}
\usepackage{xspace}
\usepackage{pifont}
\newcommand{\cmark}{\ding{51}}

\def\method{BoltNet\xspace}

\usepackage[accsupp]{axessibility}
\usepackage{hyperref}
\usepackage{orcidlink}

\begin{document}

\title{\method: An Ultra-Lightweight Convolutional Network for On-Device Plant Species Identification}
\titlerunning{\method for Embedded Plant Identification}

\author{Daniel Rossi\orcidlink{0009-0005-7774-6315} \and
Guido Borghi\orcidlink{0000-0003-2441-7524} \and
Roberto Vezzani\orcidlink{0000-0002-1046-6870}}

\authorrunning{D.~Rossi et al.}

% Springer Heidelberg, Tiergartenstr.~17, 69121 Heidelberg, Germany

\institute{
% AImageLab, Department of Engineering "Enzo Ferrari", \and
% Department of Education and Humanities, \\
University of Modena and Reggio Emilia, Modena, Italy \\
\email{\{name.surname\}@unimore.it}}

\maketitle

\begin{abstract}
Automated plant species identification from citizen-science imagery is an established, demanding fine-grained recognition problem: large taxonomic label spaces, visually similar species, and long-tailed observations require real model capacity, while field use constrains memory, latency, and power. Model size is only part of the deployment cost: intermediate activations held in memory during inference and platform-dependent execution behavior matter too, so compact recognition must be assessed on target hardware rather than through complexity metrics alone.
We present BoltNet, an ultra-lightweight fully convolutional architecture combining a Spatial Redistribution Bottleneck and Logit Pre-Sampling to improve the tradeoff between predictive performance and model size in high-cardinality classification, and report the Accuracy-Compression Tradeoff as a complementary diagnostic.
On Pl@ntNet-300K, BoltNet reaches $0.682$ F1-score with $341$K parameters ($1.37$\,MB), the highest F1-score among evaluated models below $2$\,MB and close to substantially larger convolutional backbones. Model-only measurements on a Raspberry Pi 5, Jetson Orin Nano, and Hailo-8 characterize execution across CPU, GPU, and NPU platforms, where BoltNet is the most consistently efficient model, with the best FPS/W on the GPU and NPU and second-best on the CPU. Results on AIDERv2 and CLRS provide secondary evidence of transfer across environmental image-classification tasks. Code available at: \href{https://codeberg.org/danielrossi/BoltNet}{https://codeberg.org/danielrossi/BoltNet}.


\end{abstract}
\section{Introduction}
\label{sec:intro}

Identifying plant species from field photographs underpins a growing set of ecological and agricultural applications, from citizen-science biodiversity monitoring to crop and phenotyping pipelines that increasingly rely on image analysis~\cite{joly2014interactive,chandra2020computer,remelgado2026narrowing}. The problem is well studied yet far from solved: systematic reviews describe automated species identification as a recognition task whose difficulty stems from the acquisition process itself, since the visible organ, the phenological stage, the viewpoint, and the image composition each change the evidence available to a classifier~\cite{waldchen2018plant,waldchen2018machine,seeland2017plant,rzanny2019flowers}. It is fine-grained in the strict sense, with related species differing by subtle traits while images of one species vary widely, which separates it from the balanced, low-cardinality benchmarks on which compact models are usually tuned.

Citizen-science platforms have turned this into a large-scale recognition problem. Pl@ntNet-300K assembles more than $300{,}000$ observations across $1{,}081$ species with high label ambiguity and a steep long-tailed distribution, the least represented $80\%$ of species accounting for only $11\%$ of the images~\cite{garcin2021pl}, and crowdsourced records of this kind now reach species and regions that conventional surveys cannot cover~\cite{lusk2026crowdsourced}. The long tail is not a mere statistical nuisance: rare taxa are often the most consequential to recognize, as in the early detection of invasive or emerging threats~\cite{mastin2020optimising}, and they are exactly where compact models tend to fail. The practical relevance of accurate yet small models is reinforced by evidence that freely available identification apps already reach useful accuracy for everyday users~\cite{rzanny2024more}.

Running this recognition in the field changes the binding constraint from accuracy alone to the resources of the device that carries the model, where memory, latency, and power are all limited. Model size is only one part of that cost. At inference, the peak working memory is governed by the intermediate activation tensors, which in standard convolutional networks concentrate in the early high-resolution stages and can exceed the weight footprint~\cite{lin2021memory}, while parameter count and FLOPS are themselves weak predictors of latency and energy once data movement and operator support are accounted for~\cite{sze2017efficient,ma2018shufflenet}. A model is deployable only when its weights, its activation footprint, and its operators jointly fit the target, which is why a dedicated family of ultra-lightweight networks such as EmergencyNet and TakuNet has emerged for the tightest on-device budgets, roughly an order of magnitude smaller than common lightweight and mobile backbones~\cite{kyrkou2020emergencynet,Rossi_2025_WACV}. We address this ultra-lightweight regime, which we delimit by a model size below $2$\,MB and a budget below $100$\,M FLOPS, and we benchmark against representatives of both the established lightweight backbones and the dedicated ultra-lightweight models, so that the comparison spans the spectrum rather than a single side of it.

Within this setting we present \method, an ultra-lightweight fully convolutional architecture built on two parameter-free rearrangements: a Spatial Redistribution Bottleneck in the backbone and Logit Pre-Sampling before the classifier, which together lower parameter cost while keeping dense, hardware-friendly operators (Sect.~\ref{sec:method}). On Pl@ntNet-300K, \method\ is the most accurate model below $2$\,MB and stays close to substantially larger convolutional backbones, and across a CPU, a GPU, and an NPU it is the most consistently efficient of the models we compare. AIDERv2~\cite{shianios2023benchmark} and CLRS~\cite{li2020clrs} serve as secondary transfer checks, and we report the Accuracy-Compression Tradeoff (ACT) as a complementary diagnostic of accuracy retention against parameter reduction.

The contributions of this work are:
\begin{enumerate}
\item We introduce \method, an ultra-lightweight convolutional architecture that combines the Spatial Redistribution Bottleneck and Logit Pre-Sampling to improve the recognition-to-size tradeoff in high-cardinality classification.

\item We evaluate \method\ on Pl@ntNet-300K, where it attains the highest $F_{1}$-score among the models below $2$\,MB while remaining competitive with substantially larger convolutional backbones, with secondary transfer checks on AIDERv2 and CLRS.

\item We report model-only execution across CPU, GPU, and NPU platforms, where \method\ achieves the highest FPS/W on the Jetson Orin Nano and Hailo-8 and the second-highest on the Raspberry Pi 5.

\item We report ACT as a complementary metric that jointly summarizes accuracy retention and parameter reduction across architectural variants.

\end{enumerate}

\section{Related Work}
\label{sec:related}

Our work sits at the intersection of two research directions that are usually pursued separately: image-based plant species identification under realistic acquisition conditions, and efficient visual recognition on embedded hardware. The first defines the recognition challenge, including fine-grained visual differences, heterogeneous observations, and large, imbalanced taxonomic spaces. The second concerns the architectural and system constraints that decide whether a model can run within limited memory, latency, and power budgets. We review the two directions before positioning \method\ among deployment-aware compact architectures.

\subsection{Plant identification in realistic conditions}

Image-based plant identification has moved from curated specimens and isolated organs toward photographs taken by non-experts in the field, a transition documented since the earliest social-image systems~\cite{joly2014interactive,waldchen2018plant}. This change turned the acquisition process into part of the recognition problem: depending on the observation, a classifier may receive a flower, a leaf, a fruit, a partial view, or a cluttered scene in which only a few diagnostic traits are visible. Comparative studies show that the visible organ and the image composition materially affect which cues are available, so this variability is intrinsic to the data rather than an artefact of any single model~\cite{seeland2017plant,rzanny2019flowers}. The problem is also taxonomic in nature: recognition can be assessed at genus or family level, and accuracy drops for unseen or rarely observed species~\cite{seeland2019image,waldchen2018machine}, a regime that tends to penalize low-capacity models.

Citizen-science platforms scaled these difficulties by orders of magnitude. Pl@ntNet-300K combines a $1{,}081$-class taxonomic space with strong label ambiguity and a steep long-tailed distribution, and is demanding for reasons that go beyond image count~\cite{garcin2021pl}; the structured collection of field observations that feeds such platforms is itself an active subject of study~\cite{boho2020flora}. Fine-grained plant recognition has accordingly been treated as a problem in its own right, often through specialized or transfer-learned networks~\cite{yang2022plantnet}, and recent evidence that free identification apps can reach high accuracy in everyday use makes the deployability of these models a concrete concern rather than a benchmark abstraction~\cite{rzanny2024more}. Species identification is not equivalent to plant phenotyping, but the two share field imagery, sensing platforms, and computational infrastructure, so an efficient recognition backbone can support broader plant-analysis pipelines without itself constituting a complete phenotyping solution~\cite{katal2022deep}.

\subsection{Embedded inference for environmental vision}

Field-deployed vision operates under constraints that differ from those of server-side recognition. Acquisition, inference, communication, and application logic share a single memory and power budget, and network weights must remain resident throughout execution, which is why work on memory-constrained deployment has treated model storage as a primary design variable~\cite{rusci2020memory,capotondi2020cmix}. Storage is not the only memory cost. At inference, the peak working memory is set by the largest intermediate activations, which standard convolutional designs concentrate in the early, high-resolution stages and which can dominate the weights~\cite{lin2021memory}, while the final linear layer adds a further, class-dependent burden that grows with the taxonomic space. Reducing weights therefore does not by itself bound the memory an inference allocates, but it does leave headroom for the activation working set and for the rest of the pipeline.

Deployment cost is likewise only partly captured by parameter count and FLOPS. Data movement, the memory hierarchy, tensor shapes, kernel implementation, and runtime scheduling all influence latency and energy~\cite{sze2017efficient}, and ShuffleNetV2 showed that networks with similar arithmetic complexity can differ sharply in execution time because memory-access cost and parallelism depend on the architecture and the platform~\cite{ma2018shufflenet}. Deployment frameworks such as DORY make the same point from the systems side, where practical execution hinges on compilation, tiling, and DMA scheduling~\cite{burrello2021dory}. These effects matter most when a single model is expected to run across CPUs, embedded GPUs, and dedicated NPUs, each exposing a different balance of parallelism, bandwidth, and operator support, so direct measurement on the target devices is needed to complement size and FLOPS. Compact environmental classifiers such as EmergencyNet and TakuNet show that useful recognition is possible within very small budgets, yet their primary tasks contain few classes and do not impose the fine-grained burden of a benchmark such as Pl@ntNet-300K~\cite{kyrkou2020emergencynet,Rossi_2025_WACV}.

\subsection{Embedded and ultra-lightweight architectures}

Mobile CNN research has produced a sequence of strategies for trading cost against accuracy. MobileNet introduced depthwise-separable convolutions and then the inverted residual bottleneck, in which channel expansion and linear projection balance capacity and computation~\cite{howard2017mobilenets,sandler2018mobilenetv2}; MobileNetV3 added hardware-aware search and measured latency~\cite{howard2019searching}; GhostNet reduced feature redundancy through cheap linear operations~\cite{han2020ghostnet}; ShuffleNetV2 foregrounded practical execution characteristics; and EfficientNet studied the coordinated scaling of depth, width, and input resolution~\cite{ma2018shufflenet,tan2019efficientnet,tan2021efficientnetv2}. More recent compact models incorporate attention: MobileViT and its separable-attention variant combine local convolution with global context, and EfficientFormer adapts transformer-style processing to latency-constrained deployment~\cite{mehta2021mobilevit,mehta2022separable,li2022efficientformer}. These hybrids are strong baselines rather than a foil: attention is not impractical in principle, but its operators tend to map less favorably onto the embedded runtimes considered here, so a low parameter count does not by itself identify the best deployment point, as our measurements confirm (Sect.~\ref{results}).

Taken together, the two sides of the problem are usually studied apart. Plant-identification research examines realistic observations, large taxonomic spaces, and long-tailed data, but rarely treats heterogeneous embedded execution as a primary objective; conversely, ultra-lightweight environmental models and generic mobile backbones are typically evaluated on lower-cardinality tasks or on a narrow set of devices. \method\ is positioned at this intersection. It extends the inverted residual design with the Spatial Redistribution Bottleneck, which trades channels for spatial resolution through a lossless rearrangement and so lowers the backbone's parameter cost while keeping dense kernels, in contrast to efficiency obtained through grouped, atrous, or otherwise sparse operators. Logit Pre-Sampling applies a further feature tensor shape reorganization to contain the classifier in high-cardinality settings. Pl@ntNet-300K tests whether this design preserves fine-grained recognition within an ultra-lightweight budget, measurements on CPU, GPU, and NPU assess whether the resulting operating point holds across heterogeneous runtimes, and AIDERv2 and CLRS serve as secondary checks on other environmental image-classification tasks.

\section{BoltNet}
\label{sec:method}
\method\ is a small, fully convolutional network for real-time inference on constrained edge hardware. It rests on two components: the Spatial Redistribution Bottleneck (SRB, Sect.~\ref{sec:srb}), which shapes the backbone, and Logit Pre-Sampling (LPS, Sect.~\ref{sec:lps}), which holds down the classifier.

\subsection{Spatial Redistribution Bottleneck (SRB)}
\label{sec:srb}
\begin{figure}[t]
    \centering
    \includegraphics[width=0.8\textwidth]{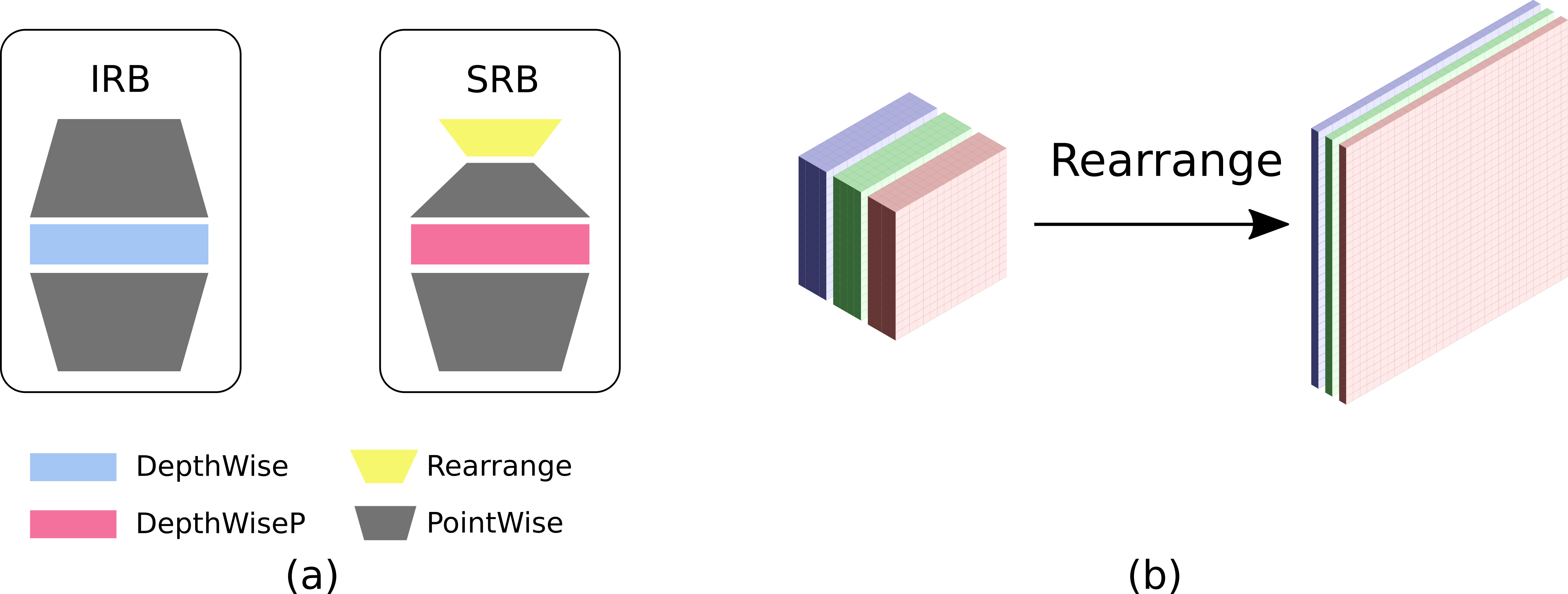}
\caption{Architecture and channel-to-spatial redistribution. (a) IRB versus SRB: the SRB adds the redistribution operation and a depthwise pooling layer (DepthWiseP) to keep spatial alignment for the skip connection. (b) The redistribution systematically moves channel-wise features into spatial locations.}
    \label{fig:architettura}
\end{figure}

A convolutional network's parameters grow with its channels, and most of that growth lands in the pointwise convolutions that mix information across channels. The inverted residual bottleneck (IRB)~\cite{sandler2018mobilenetv2} works well on mobile devices but, leaning on pointwise convolutions and channel expansion, becomes parameter-heavy in the wide later stages. The SRB attacks this by shifting part of the channel content into the spatial domain before the bottleneck~(Fig.~\ref{fig:architettura}a): a subset of channels is deterministically moved into a larger spatial grid, so the pointwise convolutions that follow see fewer channels, with nothing compressed or discarded.

Formally, let $X \in \mathbb{R}^{H \times W \times C}$ be partitioned along the channel dimension into $N=C/G$ groups $X=[X_1,\dots,X_N]$, with each $X_i \in \mathbb{R}^{H \times W \times G}$. We define a deterministic operator $f$ acting on tensor indices, inducing a bijection between the index sets $\{1,\dots,H\}\!\times\!\{1,\dots,W\}\!\times\!\{1,\dots,G\}$ and $\{1,\dots,s_h H\}\!\times\!\{1,\dots,s_w W\}\!\times\!\{1,\dots,G/(s_h s_w)\}$, where $s_h,s_w \in \mathbb{N}$ are spatial scaling factors. By construction the operator preserves cardinality, $HWG = (s_h H)(s_w W)\bigl(G/(s_h s_w)\bigr)$, performing a structured channel-to-spatial redistribution with no aggregation or information loss. Applying $f$ independently to each group gives $X'=\big\Vert_{i=1}^{N} f(X_i)\in\mathbb{R}^{\tilde{H}\times\tilde{W}\times\tilde{C}}$, with $\tilde{H}=s_h H$, $\tilde{W}=s_w W$, and $\tilde{C}=N\,G/(s_h s_w)$.

We instantiate \textit{f} with a parameter-free sub-pixel rearrangement~\cite{shi2016real}, an efficient and lossless realization of the bijection defined above. The contribution is architectural: it uses channel-to-spatial redistribution as an additional scaling axis inside the residual bottleneck, a role that does not depend on the particular rearrangement chosen. With an upscale factor of $2$ ($s_h=s_w=2$) the channel count drops fourfold while each spatial side doubles. The block then runs a $1\times1$ pointwise convolution with a nonlinearity, a $5\times5$ depthwise convolution at stride $2$ that reads the enlarged grid and brings the resolution back into line for the skip connection, and a final $1\times1$ projection.

Trading channels for spatial resolution cuts parameters and FLOPS without compressing the underlying semantics. The wider grid also gives the depthwise convolution more spatial context to work with, so it picks out finer local detail, and it nudges the feature maps into complementary roles, some tracking local structure and others holding higher-level semantics. The effect is a second scaling axis: instead of stacking depth or widening channels, an SRB network gains accuracy through denser use of the budget it has, which is what lets it push past the roughly $100$K-parameter ceiling of ultra-lightweight models while staying dense and more efficient than a plain IRB (Sect.~\ref{results}).

\subsection{\method\ Architecture}
\label{sec:arch}

\method\ consists of a convolutional stem, four progressive stages, and a classification head. Given an input of spatial size $H \times W$, the stem applies a $3 \times 3$ convolution with $32$ output channels, followed by a depthwise convolution. Both layers use stride $2$ and are followed by batch normalization and HardSwish activation, reducing the feature-map resolution to $H/4 \times W/4$ before the main network stages.

The four stages contain $4$, $4$, $4$, and $3$ SRB blocks with output widths of $24$, $56$, $152$, and $368$ channels, respectively. A max-pooling layer is applied at the end of each stage, except for the last one, to progressively reduce the spatial resolution while increasing the channel capacity.

The classification head applies Logit Pre-Sampling to the final feature tensor, followed by global average pooling and a linear classifier. HardSwish~\cite{howard2019searching} is used throughout the network except for the output layer, and batch normalization follows every convolutional operation.

\subsection{Logit Pre-Sampling (LPS)}
\label{sec:lps}
In traditional image classification architectures, the final linear classifier can account for a substantial fraction of the total parameter budget, and this imbalance grows more pronounced as the number of classes increases. Given a pre-logit representation with $C$ channels and a classification problem with $N_{\mathrm{cls}}$ classes, a conventional linear head contains $C N_{\mathrm{cls}}$ weights. On Pl@ntNet-300K, where $N_{\mathrm{cls}}=1{,}081$, the classifier can therefore become disproportionately large relative to an ultra-lightweight backbone.

Logit Pre-Sampling (LPS) reduces this cost by applying a parameter-free channel-to-spatial rearrangement to the final feature tensor before global average pooling. Let $\mathbf{X} \in \mathbb{R}^{H \times W \times C}$ denote the pre-logit feature map, and let $r$ be the LPS sampling factor, with $C$ divisible by $r^{2}$. LPS applies a rearrangement operator $\mathcal{P}_{r}$ \cite{shi2016real},
\[
\mathbf{X}' = \mathcal{P}_{r}(\mathbf{X}), \qquad \mathbf{X}' \in \mathbb{R}^{rH \times rW \times C/r^{2}},
\]
which is bijective and discards no feature values, changing only their organization across the channel and spatial dimensions.

Global average pooling is then applied to the rearranged tensor,
\[
\mathbf{z} = \operatorname{GAP}(\mathbf{X}') \in \mathbb{R}^{C/r^{2}},
\]
and the classifier is
\[
\mathbf{y} = \mathbf{W}\mathbf{z}+\mathbf{b}, \qquad \mathbf{W} \in \mathbb{R}^{N_{\mathrm{cls}} \times C/r^{2}}.
\]
Ignoring the bias term, LPS reduces the classifier parameter count from $C N_{\mathrm{cls}}$ to $C N_{\mathrm{cls}}/r^{2}$. The reduction therefore grows with both the sampling factor and the number of output classes, making LPS particularly relevant to high-cardinality recognition tasks.

The rearrangement itself is lossless, while the subsequent global pooling performs a structured aggregation of groups of pre-logit channels. LPS does not alter the convolutional backbone and introduces no learnable parameters. Its effect on model size and predictive performance is evaluated independently in Sect.~\ref{results}.

\section{Experiments}
\label{sec:experiments}

\subsection{Benchmark Datasets}
\label{sec:datasets}

The evaluation spans three complementary environmental recognition settings. Pl@ntNet-300K is the primary benchmark and targets species-level identification from citizen-science imagery. AIDERv2 and CLRS provide secondary evaluations on aerial disaster recognition and remote-sensing scene classification, respectively, allowing us to assess whether the proposed architecture remains effective across different image domains, label-space sizes, and acquisition conditions.

\paragraph{Pl@ntNet-300K~\cite{garcin2021pl}.}
Pl@ntNet-300K is derived from observations submitted to the Pl@ntNet citizen-science platform and contains $306{,}146$ images covering $1{,}081$ plant species from $303$ genera. The images are collected under uncontrolled field conditions and vary considerably in viewpoint, background, scale, composition, and visible plant organs. The dataset also exhibits substantial label ambiguity and a strongly long-tailed distribution, with the least represented $80\%$ of species accounting for only $11\%$ of the images. These characteristics make it a demanding benchmark for fine-grained species recognition and particularly suitable for evaluating whether an ultra-lightweight model can retain sufficient discriminative capacity over a large taxonomic label space. We use the official split of $243{,}916$ training, $31{,}118$ validation, and $31{,}112$ test images.

\paragraph{AIDERv2~\cite{shianios2023benchmark}.}
AIDERv2 is an aerial-image benchmark developed for disaster recognition and emergency monitoring. It contains approximately $16{,}000$ multi-resolution images assigned to four classes: collapsed buildings, fires, floods, and normal scenes. The images originate from heterogeneous aerial sources and cover a range of environments and acquisition conditions, including variations in viewpoint, scene scale, resolution, and background content. From an environmental perspective, the dataset represents event-level recognition, where a model must distinguish hazardous situations from normal conditions in imagery that may be acquired by UAVs during rapid situational assessment. We use the standard $80\%/10\%/10\%$ training, validation, and test split.

\paragraph{CLRS~\cite{li2020clrs}.}
CLRS is a remote-sensing scene-classification dataset containing $15{,}000$ RGB images uniformly distributed across $25$ categories, with $600$ images per class. Each image has a spatial resolution of $256 \times 256$ pixels, while the metric resolution of the images ranges from $0.26m$ to $8.85m$. Its categories cover a broad set of land-cover and land-use patterns, including natural, agricultural, urban, and infrastructure environments. CLRS therefore complements the object-level and event-level settings of Pl@ntNet-300K and AIDERv2 with landscape-level visual interpretation from overhead imagery. We use the dataset as a balanced secondary benchmark and adopt a $70\%/10\%/20\%$ training, validation, and test split.

\subsection{Training Settings}
\label{subsec:settings}

All images are normalized and resized to $224 \times 224$ pixels. For Pl@ntNet-300K, we apply a restrained combination of geometric, noise, and photometric transformations to preserve fine-grained visual characteristics. For AIDERv2 and CLRS, we follow the augmentation protocol introduced by Kyrkou \etal~\cite{kyrkou2020emergencynet}.

\method\ is trained from scratch for $300$ epochs using SGD with momentum $0.9$ and a batch size of $256$. We minimize cross-entropy loss with label smoothing of $0.1$ and weight decay of $5 \times 10^{-5}$. The learning rate follows a cosine-annealing schedule from an initial value of $\eta_{\max}=0.05$ to $\eta_{\min}=8 \times 10^{-5}$. All experiments use random seed $22$.

For the model comparison, every architecture is trained from randomly initialized weights on the target dataset. Each baseline retains the architecture-specific optimization recipe reported in its original publication, while the dataset split, input resolution, augmentation pipeline, and random seed are held fixed across models.

\subsection{Evaluation Protocol and Hardware}
\label{subsec:protocol}

We evaluate predictive performance using the weighted $F_1$ score, which is more informative than accuracy in the presence of class imbalance. Model complexity is reported in terms of parameter count and FLOPS. However, since these theoretical metrics do not reliably predict deployment cost~\cite{ma2018shufflenet}, our analysis focuses primarily on measurements collected on the target hardware. For each model--platform pair, we report mean inference throughput in frames per second (FPS), average power consumption in watts (W), and energy efficiency as FPS/W.

The evaluated architectures cover three model families: mobile CNNs, including MobileNet, EfficientNet, and RegNet variants; ultra-lightweight edge models, including EmergencyNet and TakuNet; and convolution--attention hybrids, including MobileViT and EfficientFormer variants. Mobile and ultra-lightweight architectures are analyzed separately because the latter satisfy substantially tighter complexity constraints, with model sizes below $2$\,MB and computational costs below $100$\,M FLOPS.

\noindent\textbf{Hardware.}
To assess whether model efficiency is preserved across heterogeneous deployment backends, we benchmark all compatible models on three representative edge platforms: a Raspberry~Pi~5 with an ARM Cortex-A76 CPU, an NVIDIA Jetson Orin Nano 8G with an Ampere GPU, and a Hailo-8 dataflow NPU. These platforms represent general-purpose CPU execution, GPU-based parallel acceleration, and dedicated neural-network acceleration, respectively. Evaluating the same model families across these targets provides a more realistic assessment of deployment efficiency than hardware-independent complexity metrics alone.

\subsection{Accuracy-Compression Tradeoff (ACT)}
\label{subsec:act}
As a complementary read on how gracefully a model trades accuracy for size, we report the Accuracy-Compression Tradeoff (ACT), a dimensionless score that penalizes accuracy loss non-linearly and rewards parameter reduction with diminishing returns:
\begin{equation}
\label{eq:act}
ACT = \left( \frac{\text{acc}_{\text{comp}}}{\text{acc}_{\text{orig}}} \right)^{k}
      \cdot \log_2\!\left( 1 + \frac{p_{\text{orig}}}{p_{\text{comp}}} \right),
\end{equation}
where $\text{acc}$ is the metric and $p$ the parameter count, with subscripts for the original and compressed models. The fidelity term $(\text{acc}_{\text{comp}}/\text{acc}_{\text{orig}})^{k}$ makes accuracy loss bite geometrically, the exponent $k$ setting how strongly fidelity outweighs size; the efficiency term $\log_2(1+p_{\text{orig}}/p_{\text{comp}})$ measures the gain from compression, and the unit shift keeps it non-negative. We interpret ACT only as a diagnostic metric. It is designed exclusively to identify, within the same model family, the architecture that most effectively compresses predictive performance into its parameter budget. Consequently, we do not use it as a general criterion for comparing different architectures or selecting the best overall model, but rather as a tool to better understand the effects of architectural design choices, alongside F\textsubscript{1} and the on-device measurements.

\section{Results}
\label{results}
% We report efficiency before accuracy, because for a field system the first question is whether a model runs at all within the board's budget. Throughout, we keep the dedicated ultra-lightweight models (EmergencyNet, TakuNet, and \method) separate from the larger lightweight and attention backbones. The two groups differ by roughly an order of magnitude in size, and reading them together would let raw capacity obscure the size-constrained comparison that decides deployment.

\subsection{On-device efficiency}
Table~\ref{tab:embedded_benchmark} reports end-to-end inference on the three boards, and the ranking shifts with the execution model. \method\ takes the best energy efficiency on the Jetson GPU ($30.7$ FPS/W) and the Hailo NPU ($2354$ FPS/W), and is a close second on the Raspberry~Pi~CPU ($8.9$ against TakuNet's $9.2$). Each competitor marks a different limit. TakuNet wins the CPU, where parallelism is scarce, but its sparse blocks underuse wide accelerators, so its NPU and GPU efficiency fall to roughly $0.4$ and $0.7$ times \method's at comparable model size. RegNet, a wide dense backbone, posts the highest raw NPU frame rate ($5795$ FPS) because regular convolutions map cleanly onto the dataflow array, yet it sustains this at $2.5$\,W against \method's $1.6$\,W, so the two reach almost the same FPS/W with \method\ drawing about a third less power. EmergencyNet falls in between: its atrous depthwise fusion enlarges the receptive field cheaply but scatters memory access, keeping its efficiency below \method\ on every board. MobileViT~V2 is the slowest and least efficient model everywhere, about $19\times$ below \method\ on the NPU, because attention's matrix products and reshapes still lack efficient edge kernels. \method\ is the only network that stays at or near the efficiency frontier on all three execution models, a desirable aspect for a single deployable backbone.

These differences follow arithmetic intensity rather than nominal cost. The SRB narrows channels through a lossless rearrangement instead of grouped, atrous, or sparse operators, so the work that remains is standard dense pointwise and depthwise convolution, which keeps the accelerators busy. Designs that save FLOPS through irregular operators instead issue many small, memory-bound kernels that stall on parallel hardware, which is why TakuNet leads the scalar CPU but neither the GPU nor the NPU. That a model of TakuNet's size can be $2.4\times$ less efficient than \method\ on the NPU is the clearest single sign that parameter count and FLOPS do not predict deployed cost, and that the on-device numbers are the ones to trust.

\begin{table}[t]
\centering
\setlength{\tabcolsep}{3pt}
\begin{tabular}{l*{9}{c}}
\toprule
\multirow{3}{*}{\textbf{Model}} & \multicolumn{3}{c}{\textbf{Raspberry Pi 5}} & \multicolumn{3}{c}{\textbf{Hailo 8}} & \multicolumn{3}{c}{\textbf{Jetson Orin Nano}} \\
 & \multicolumn{3}{c}{\textit{(CPU - Cortex A76)}} & \multicolumn{3}{c}{\textit{(NPU - Hailo Arch)}} & \multicolumn{3}{c}{\textit{(GPU - Tegra Ampere)}} \\
\cmidrule(lr){2-4} \cmidrule(lr){5-7} \cmidrule(lr){8-10}
 & {\textbf{FPS}} & {\textbf{W}} & {\textbf{FPS/W}} & {\textbf{FPS}} & {\textbf{W}} & {\textbf{FPS/W}} & {\textbf{FPS}} & {\textbf{W}} & {\textbf{FPS/W}} \\
\midrule
MobileVitV2 & 27.9 & 9.2 & 3.0 & 130.3 & 1.1 & 123.4 & 148.0 & 11.7 & 12.7 \\
RegNet      & 72.2 & 9.3 & 7.7 & \textbf{5794.7} & 2.5 & \underline{2317.9} & \underline{302.6} & 12.0 & \underline{25.2} \\
EmergencyNet& 79.8 & 9.1 & 8.8 & 1698.1 & 1.8 & 962.6 & 284.3 & 11.9 & 23.9 \\
TakuNet     & \textbf{104.6} & 11.4 & \textbf{9.2} & 1440.4 & 1.5 & 988.6 & 252.4 & 11.4 & 22.1 \\
\textbf{\method} & \underline{94.0} & 10.6 & \underline{8.9} & \underline{3778.8} & 1.6 & \textbf{2354.4} & \textbf{325.4} & 10.6 & \textbf{30.7} \\
\bottomrule
\end{tabular}
\caption{Inference benchmarks on three edge platforms of models trained on Pl@ntNet300K, with \textbf{best} and \underline{second best} highlighted.}
\label{tab:embedded_benchmark}
\end{table}

\subsection{Accuracy on long-tailed species recognition}
On Pl@ntNet-300K (Table~\ref{tab:PlantNet300K_results}) \method\ reaches $0.682$ F1-score at $0.34$\,M parameters, the best of the deployable models below $2$\,MB: ahead of EmergencyNet ($0.636$) with $8\%$ fewer parameters and half the FLOPS, and far ahead of TakuNet ($0.483$). Two design choices account for the margin over the other ultra-lightweight models. The SRB relocates capacity into the spatial grid rather than removing it, so a small budget stays usable, and Logit Pre-Sampling keeps the head from consuming that budget: on $1{,}081$ classes a plain classifier over the final $368$ channels would need about $0.40$\,M weights, more than the whole network, and LPS cuts this roughly fourfold. The ablation makes the effect concrete, with the head accounting for the $0.30$\,M-parameter gap between the SRB-only model and \method. The earlier ultra-lightweight networks have neither mechanism, so their already small backbones are further starved by an oversized head. Against larger networks \method\ is competitive rather than dominant: it trails RegNet ($0.702$) by two F1-score points at one-eighth the parameters, slightly exceeds FBNetV3 ($0.678$) at one twenty-fifth of its size, and is above MobileNetV2, MobileNetV3, EfficientNet-B0, and EfficientNetV2-S, several of which do not turn their nominal capacity into accuracy when trained from scratch on this long tail dataset (Sect.~\ref{subsec:settings}). The attention models are the most accurate, MobileViT at $0.741$ and $0.732$ and EfficientFormer at $0.714$, since a global receptive field aids fine-grained discrimination, but Table~\ref{tab:embedded_benchmark} places them last for deployment by a wide margin. The benchmark is what gives the result weight: its label ambiguity and steep long tail penalize low-capacity models~\cite{garcin2021pl}, which is why the prior ultra-lightweight networks fall so far short of \method\ here.

\begin{table}[t]
\centering
\begin{tabular}{l c c c c c}
\toprule
\textbf{Model} & \textbf{Parameters} & \textbf{Model Size (MB)} & \textbf{F\textsubscript{1}-score} & \textbf{FLOPS (G)} \\
\midrule
EfficientNetV2 S \cite{tan2021efficientnetv2}& 21,562,249 & 85.63 & 0.443 & 2.874 \\
FBNetV3 \cite{dai2021fbnetv3}& 8,759,249 & 34.85 & 0.678 & 0.422 \\
MobileNetV3 L-100 \cite{howard2019searching} & 5,586,793 & 22.25 &  0.434 & \underline{0.226} \\
EfficientNet B0 \cite{tan2019efficientnet} & 5,392,309 & 21.40 & 0.493 & 0.399 \\
MobileNetV2 100 \cite{sandler2018mobilenetv2} & 3,608,633 & 14.30 & 0.444 & 0.314 \\
EfficientFormerV2 S \cite{li2023rethinking} & 3,628,930 & 14.42 & 0.714 & 0.407 \\
RegNetX 002\cite{radosavovic2020designing} & 2,714,681 & 10.78 & 0.702 & \textbf{0.203} \\
MobileVitV2 050 \cite{mehta2022separable}& \underline{1,391,410} & \underline{5.53} & \textbf{0.741} & 0.374 \\
MobileVit XXS \cite{mehta2021mobilevit}& \textbf{1,298,025} & \textbf{5.18} & \underline{0.732} & 0.263 \\
\midrule
EmergencyNet \cite{kyrkou2020emergencynet} & 369,647 & 1.48 & \underline{0.636} & 0.116 \\
TakuNet \cite{Rossi_2025_WACV} & \textbf{297,001} & \textbf{1.19} & 0.483 & \textbf{0.032} \\
\method & \underline{341,254} & \underline{1.37} & \textbf{0.682} & \underline{0.056} \\
\bottomrule
\end{tabular}
\caption{Pl@ntNet-300K~\cite{garcin2021pl}: \textbf{best} and \underline{second-best} per group.}
\label{tab:PlantNet300K_results}
\end{table}

\begin{table}[th!]
\centering
\begin{tabular}{l c c c c}
\toprule
\textbf{Model} & \textbf{Parameters} & \textbf{Model Size (MB)} & \textbf{F\textsubscript{1}-score} & \textbf{FLOPS (G)}\\
\midrule
EfficientNetV2 s\cite{tan2021efficientnetv2} & 20,182,612 & 80.11 & 0.817 & 2.873 \\
FBNetV3 \cite{dai2021fbnetv3}& 6,621,404 & 26.30 & 0.924 & 0.419 \\
MobileNetV3 L-100\cite{howard2019searching} & 4,207,156 & 16.73 & 0.961 & \underline{0.224} \\
EfficientNet-B0 \cite{tan2019efficientnet} & 4,012,672 & 15.88 & 0.949 & 0.398 \\
EfficientFormerV2 S \cite{li2023rethinking} & 3,247,672 & 12.90 & \textbf{0.968} & 0.407 \\
RegNetX 002\cite{radosavovic2020designing} & 2,317,268 & 9.19 & 0.946 & \textbf{0.203} \\
MobileNetV2 \cite{sandler2018mobilenetv2} & 2,228,996 & 8.78 & 0.956 & 0.313 \\
MobileViT V2 050 \cite{mehta2022separable} & \underline{1,114,621} & \underline{4.43} & \underline{0.966} & 0.374 \\
MobileVit XXS \cite{mehta2021mobilevit} & \textbf{952,308} & \textbf{3.79} & 0.962 & 0.263 \\
\midrule
EmergencyNet \cite{kyrkou2020emergencynet} & \underline{90,704} & \underline{0.36} & 0.952 & 0.062 \\
TakuNet\cite{Rossi_2025_WACV} & \textbf{37,444} & \textbf{0.15} & \underline{0.953} & \textbf{0.031} \\
\method &  241,093 & 0.96 & \textbf{0.958} & \underline{0.055} \\
\bottomrule
\end{tabular}
\caption{AIDERv2~\cite{shianios2023benchmark}: \textbf{best} and \underline{second-best} per group.}
\label{tab:AIDERV2_results}
\end{table}

\subsection{Cross-domain generalization under scarce data}
Tables~\ref{tab:AIDERV2_results} and~\ref{tab:CLRS_results} are transfer checks, not a second contribution: they ask whether a design tuned for plant recognition still behaves well under domain shift and on the smaller environmental datasets common at the edge. On AIDERv2 the few classes and limited data saturate most models within a narrow band, and inside it \method\ is the best deployable convolutional network ($0.958$ F1); the enlarged spatial grid from the SRB plausibly helps with the small local structures typical of aerial views. The harder, higher-cardinality CLRS is more discriminating: \method\ reaches $0.825$, on par with MobileViT~V2 ($0.826$) at about one-fifth of the parameters and one-seventh of the FLOPS, and ahead of every ultra-lightweight rival. The behavior measured on Pl@ntNet-300K therefore carries over to aerial and remote-sensing imagery without modification, which is the evidence we ask these datasets to provide.

\begin{table}[t]
\centering
\begin{tabular}{l c c c c c}
\toprule
\textbf{Model} & \textbf{Parameters} & \textbf{Model Size (MB)} & \textbf{F\textsubscript{1}-score} & \textbf{FLOPS (G)} \\
\midrule
EfficientNetV2 S \cite{tan2021efficientnetv2}& 20,209,513 & 80.22 & 0.707 & 2.873 \\
FBNetV3 \cite{dai2021fbnetv3}& 6,663,089 & 26.47 & 0.668 & 0.420 \\
MobileNetV3 L-100 \cite{howard2019searching} & 4,234,057 & 16.84 & 0.768 & \underline{0.224} \\
EfficientNet B0 \cite{tan2019efficientnet} & 4,039,573 & 15.99 & 0.780 & 0.398 \\
EfficientFormerV2 S \cite{li2023rethinking} & 3,255,106 & 12.93 & 0.833 & 0.407 \\
RegNetX 002\cite{radosavovic2020designing} & 2,325,017 & 9.22 & \underline{0.836} & \textbf{0.203} \\
MobileNetV2 100 \cite{sandler2018mobilenetv2} & 2,255,897 & 8.89 & 0.603 & 0.313 \\
MobileVitV2 050 \cite{mehta2022separable}& \underline{1,120,018} & \underline{4.45} & 0.826 & 0.374 \\
MobileVit XXS \cite{mehta2021mobilevit}& \textbf{959,049} & \textbf{3.82} & \textbf{0.860} & 0.264 \\
\midrule
EmergencyNet \cite{kyrkou2020emergencynet} & \underline{96,143} & \underline{0.38} & \underline{0.773} & 0.063 \\
TakuNet \cite{Rossi_2025_WACV} & \textbf{42,505} & \textbf{0.17} & 0.760 & \textbf{0.031} \\
\method & 243,046 & 0.97 & \textbf{0.825} & \underline{0.055} \\
\bottomrule
\end{tabular}
\caption{CLRS~\cite{li2020clrs}: \textbf{best} and \underline{second-best} per group.}
\label{tab:CLRS_results}
\end{table}

\begin{table}[th!]
\centering
\begin{tabular}{lcccc | llll}
\toprule
 & \textbf{SRB} & \textbf{$k_{SRB}$} &  \textbf{LPS}& \textbf{$k_{LPS}$} & \textbf{Parameters} & \textbf{FLOPS} & \textbf{F\textsubscript{1}-score} & \textbf{ACT}\\
\midrule
IRBNet  &  &  &  &  & $1{,}486{,}029$ & $154.4$M & $0.704$ & $1.000$ \\ \midrule
        & \cmark & $2$ &  &  & $639{,}610$ {\scriptsize \textcolor{ForestGreen}{$-56.9$\%}} & $56.1$M {\scriptsize \textcolor{ForestGreen}{$-63.7$\%}} & $0.687$ {\scriptsize \textcolor{BrickRed}{$-2.4$\%}} & $1.460$ \\
        & \cmark & $3^{\star}$ &  &  & $541{,}930$ {\scriptsize \textcolor{ForestGreen}{$-63.5$\%}} & $40.8$M {\scriptsize \textcolor{ForestGreen}{$-73.6$\%}} & $0.628$ {\scriptsize \textcolor{BrickRed}{$-10.8$\%}} & $0.856$\\
        &  &  & \cmark & $2$ & $1{,}187{,}673$ {\scriptsize \textcolor{ForestGreen}{$-20.1$\%}} & $154.1$M {\scriptsize \textcolor{ForestGreen}{$-0.2$\%}} & $0.715$ {\scriptsize \textcolor{ForestGreen}{$+1.56$\%}} & $1.305$ \\
        &  &  & \cmark & $3^{\star}$ & $1{,}136{,}183$ {\scriptsize \textcolor{ForestGreen}{$-23.5$\%}} & $154.2$M {\scriptsize \textcolor{ForestGreen}{$-0.1$\%}} & $0.721$ {\scriptsize \textcolor{ForestGreen}{$+2.41$\%}} & $1.426$ \\
        \midrule
        \textbf{\method} & \cmark & $2$ & \cmark & $2$ & $341{,}254$ {\scriptsize \textcolor{ForestGreen}{$-77.0$\%}} & $55.8$M {\scriptsize \textcolor{ForestGreen}{$-63.9$\%}} & $0.682$ {\scriptsize \textcolor{BrickRed}{$-3.1$\%}} & $1.938$\\
\bottomrule
\end{tabular}
\caption{Ablation of \method\ on Pl@ntNet-300K. SRB = Spatial Redistribution Bottleneck, LPS = Logit Pre-Sampling, $k$ = block upscale/sampling factor. IRBNet is the inverted-bottleneck-only baseline.}
\label{tab:ablation_architecture}
\end{table}

\subsection{Ablation and compression diagnostic}
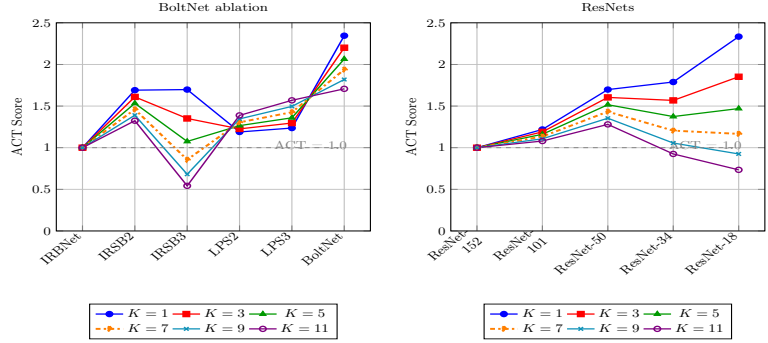
\begin{figure}[th!]
    \centering
    \makebox[\textwidth]{
        \begin{minipage}{0.4\textwidth}
            \centering
            \resizebox{\linewidth}{4.5cm}{\begin{tikzpicture}
    \begin{axis}[
        title={BoltNet ablation},
        ylabel={ACT Score},
        ymin=0, ymax=2.5,
        xtick=data, 
        symbolic x coords={IRBNet, IRSB2, IRSB3, LPS2, LPS3, BoltNet}, 
        xticklabel style={rotate=45, anchor=east, text width=1.5cm, align=right}, 
        legend style={at={(0.5,-0.35)}, anchor=north, legend columns=3},
        grid=major,
        % Modificato a \linewidth per lavorare correttamente all'interno di \minipage
    ]

    % K = 1 (Blu)
    \addplot[blue, mark=*, line width=1pt] coordinates {
        (IRBNet, 1.000) (IRSB2, 1.691) (IRSB3, 1.698) (LPS2, 1.189) (LPS3, 1.236) (BoltNet, 2.345)
    };
    \addlegendentry{$K=1$}

    % K = 3 (Rosso)
    \addplot[red, mark=square*, line width=1pt] coordinates {
        (IRBNet, 1.000) (IRSB2, 1.610) (IRSB3, 1.351) (LPS2, 1.226) (LPS3, 1.296) (BoltNet, 2.201)
    };
    \addlegendentry{$K=3$}

    % K = 5 (Verde)
    \addplot[green!60!black, mark=triangle*, line width=1pt] coordinates {
        (IRBNet, 1.000) (IRSB2, 1.533) (IRSB3, 1.075) (LPS2, 1.265) (LPS3, 1.359) (BoltNet, 2.065)
    };
    \addlegendentry{$K=5$}
    
    % K = 7 (Arancione/Oro)
    \addplot[orange, mark=diamond*, line width=1.5pt, dash dot] coordinates {
        (IRBNet, 1.000) (IRSB2, 1.460) (IRSB3, 0.856) (LPS2, 1.305) (LPS3, 1.426) (BoltNet, 1.938)
    };
    \addlegendentry{$K=7$}
    
    % K = 9 (Ciano)
    \addplot[cyan!70!black, mark=x, line width=1pt] coordinates {
        (IRBNet, 1.000) (IRSB2, 1.390) (IRSB3, 0.681) (LPS2, 1.346) (LPS3, 1.496) (BoltNet, 1.819)
    };
    \addlegendentry{$K=9$}
    
    % K = 11 (Viola)
    \addplot[violet, mark=o, line width=1pt] coordinates {
        (IRBNet, 1.000) (IRSB2, 1.324) (IRSB3, 0.542) (LPS2, 1.388) (LPS3, 1.569) (BoltNet, 1.707)
    };
    \addlegendentry{$K=11$}

    % Linea di riferimento ACT = 1.0
    \draw[dashed, black!50] (axis cs:IRBNet,1.0) -- (axis cs:BoltNet,1.0) node[anchor=north west, pos=0.8, xshift=-0.5cm, yshift=0.3cm] {ACT = 1.0};
    
    \end{axis}
\end{tikzpicture}}
        \end{minipage}
        \hspace{0.01\textwidth}
        \begin{minipage}{0.4\textwidth}
            \centering
            \resizebox{\linewidth}{4.5cm}{\begin{tikzpicture}
    \begin{axis}[
        title={ResNets},
        ylabel={ACT Score},
        ymin=0, ymax=2.5,
        xtick=data, 
        symbolic x coords={ResNet-152, ResNet-101, ResNet-50, ResNet-34, ResNet-18}, 
        xticklabel style={rotate=45, anchor=east, text width=1.5cm, align=right}, 
        legend style={at={(0.5,-0.35)}, anchor=north, legend columns=3}, 
        grid=major,
        % Larghezza corretta per l'uso in \minipage
    ]

    % K = 1 (Blu)
    \addplot[blue, mark=*, line width=1pt] coordinates {
        (ResNet-152, 1.000) (ResNet-101, 1.219) (ResNet-50, 1.698) (ResNet-34, 1.789) (ResNet-18, 2.334)
    };
    \addlegendentry{$K=1$}

    % K = 3 (Rosso)
    \addplot[red, mark=square*, line width=1pt] coordinates {
        (ResNet-152, 1.000) (ResNet-101, 1.190) (ResNet-50, 1.604) (ResNet-34, 1.568) (ResNet-18, 1.852)
    };
    \addlegendentry{$K=3$}

    % K = 5 (Verde)
    \addplot[green!60!black, mark=triangle*, line width=1pt] coordinates {
        (ResNet-152, 1.000) (ResNet-101, 1.161) (ResNet-50, 1.516) (ResNet-34, 1.374) (ResNet-18, 1.470)
    };
    \addlegendentry{$K=5$}
    
    % K = 7 (Arancione - Evidenziato)
    \addplot[orange, mark=diamond*, line width=1.5pt, dash dot] coordinates {
        (ResNet-152, 1.000) (ResNet-101, 1.134) (ResNet-50, 1.433) (ResNet-34, 1.205) (ResNet-18, 1.166)
    };
    \addlegendentry{$K=7$}
    
    % K = 9 (Ciano)
    \addplot[cyan!70!black, mark=x, line width=1pt] coordinates {
        (ResNet-152, 1.000) (ResNet-101, 1.107) (ResNet-50, 1.354) (ResNet-34, 1.056) (ResNet-18, 0.925)
    };
    \addlegendentry{$K=9$}
    
    % K = 11 (Viola)
    \addplot[violet, mark=o, line width=1pt] coordinates {
        (ResNet-152, 1.000) (ResNet-101, 1.080) (ResNet-50, 1.280) (ResNet-34, 0.925) (ResNet-18, 0.734)
    };
    \addlegendentry{$K=11$}

    % Linea di riferimento ACT = 1.0
    \draw[dashed, black!50] (axis cs:ResNet-152,1.0) -- (axis cs:ResNet-18,1.0) node[anchor=north west, pos=0.8, xshift=-0.5cm, yshift=0.3cm] {ACT = 1.0};
    
    \end{axis}
\end{tikzpicture}}
        \end{minipage}
}
\caption{Accuracy-Compression Tradeoff (ACT, $\uparrow$) for \method\ ablations and the ResNet family. As the sensitivity coefficient $k$ grows, the models with a superior accuracy-parameter tradeoff emerge.}
    \label{fig:ACT}
\end{figure}

Table~\ref{tab:ablation_architecture} separates the components on Pl@ntNet-300K against an inverted-bottleneck baseline (IRBNet). The SRB carries the backbone compression: at upscale factor $2$ it removes $56.9\%$ of the parameters and $63.7\%$ of the FLOPS for a $2.4\%$ relative drop in F1, whereas factor $3$ costs $10.8\%$, so the redistribution has a capacity floor and must be sized to the task. LPS behaves differently depending on where it acts. On the full backbone it slightly improves accuracy ($+1.6\%$ at factor $2$) while removing a fifth of the parameters, because shrinking an oversized head also regularizes it. Placed on top of the already-compressed SRB backbone, that accuracy gain no longer transfers: the step from the SRB-only model to \method\ removes the $0.30$\,M-parameter head for a $0.005$ drop in F1 ($0.687$ to $0.682$). The two components are thus complementary in what they compress, the backbone and the head, rather than additive in accuracy, and together they reach $-77\%$ parameters at $-3.1\%$ F1. The ACT curve (Fig.~\ref{fig:ACT}), with sensitivity $k=7$ fixed against established backbones such as ResNet~\cite{he2016deep}, ranks \method\ highest, consistent with a design that relocates capacity rather than discarding it. Indeed, as the sensitivity increases, accuracy retention gets stricter over a fixed compression ratio, highlighting models which effectively retain accuracy despite the reduced internal representation capacity.  Entries marked $(\star)$ adjust widths to the operator's channel-divisibility requirement, a multiple of the squared upscale factor.

% \section{Limitations}
% \label{sec:limitations}

% The compact operating point of \method\ involves a measurable recognition tradeoff. Relative to IRBNet, the complete model reduces the parameter count by $77.0\%$ and FLOPS by $63.9\%$, while the $F_{1}$-score decreases from $0.704$ to $0.682$. Larger mobile and attention-based architectures also retain higher absolute recognition performance on Pl@ntNet-300K. We consider this compromise appropriate for embedded deployment, where model weights must fit within limited persistent-memory budgets and coexist with the remaining application components~\cite{rusci2020memory,capotondi2020cmix}. \method\ should therefore be interpreted as an accuracy-to-memory operating point rather than as a replacement for larger models when recognition accuracy is the only objective.

% The ablation also shows that spatial redistribution cannot be increased arbitrarily. An upscale factor of $3$ provides additional parameter and FLOPS reduction but causes a $10.8\%$ relative decrease in $F_{1}$-score compared with IRBNet. Moreover, the redistribution operation requires compatible channel widths, which must be divisible by the squared upscale factor. The factor and stage widths must consequently be selected according to the capacity requirements of the target task.

\section{Conclusions}
\label{sec:conclusions}
We presented \method, an ultra-lightweight architecture for plant species identification on constrained field hardware. Its Spatial Redistribution Bottleneck moves channel content into the spatial grid through a parameter-free, lossless rearrangement, cutting parameters and computation while letting the depthwise convolutions resolve finer local detail, and Logit Pre-Sampling applies the same idea to keep the classifier head small. On the fine-grained, long-tailed Pl@ntNet-300K benchmark \method\ is the most accurate model in its deployable size class, and across a CPU, a GPU, and an NPU it delivers the best measured energy efficiency on the GPU and the NPU and near-best on the CPU; transfer checks on AIDERv2 and CLRS show the design holds up under domain shift and with scarce data. Rather than claiming an ecological outcome, we offer \method\ as an efficient, deployable backbone on which field identification and, in time, on-device phenotyping systems can be built.

\bibliographystyle{splncs04}
\bibliography{main}

\end{document}